\documentclass[sigconf]{acmart}
\AtBeginDocument{%
  }

\setcopyright{acmlicensed}
\copyrightyear{2025}
\acmYear{2025}
\setcopyright{acmlicensed}\acmConference[MM '25]{Proceedings of the 33rd ACM International Conference on Multimedia}{October 27--31, 2025}{Dublin, Ireland}
\acmBooktitle{Proceedings of the 33rd ACM International Conference on Multimedia (MM '25), October 27--31, 2025, Dublin, Ireland}
\acmDOI{10.1145/3746027.3755216}
\acmISBN{979-8-4007-2035-2/2025/10}
\usepackage{multirow}

\begin{document}

\title{Reliable Cross-modal Alignment via Prototype Iterative Construction}

\author{Xiang Ma}
\orcid{0000-0002-4963-8705}
\affiliation{%
	\institution{Shandong University}
	\city{Jinan, Shandong}
	\country{China}}
\email{xiangma@sdu.edu.cn}

\author{Litian Xu}
\orcid{0009-0007-5588-5642}
\affiliation{%
	\institution{The University of Exeter}
	\city{Exeter}
	\country{United Kingdom}}
\email{lx268@exeter.ac.uk}

\author{Lexin Fang}
\orcid{0000-0003-3243-851X}
\affiliation{%
	\institution{Shandong University}
	\city{Jinan, Shandong}
	\country{Chian}}
\email{fanglexin@mail.sdu.edu.cn}

\author{Caiming Zhang}
\orcid{0000-0003-0217-1543}
\affiliation{%
	\institution{Shandong University}
	\city{Jinan, Shandong}
	\country{China}}
\email{czhang@sdu.edu.cn}

\author{Lizhen Cui}
\orcid{0000-0002-8262-8883}
\authornote{Corresponding Author}
\affiliation{%
	\institution{Shandong University}
	\institution{The Joint SDU-NTU Centre for Artificial Intelligence Research}
	\city{Jinan, Shandong}
	\country{China}}
\email{clz@sdu.edu.cn}

\renewcommand{\shortauthors}{Xiang Ma, Litian Xu, Lexin Fang, Caiming Zhang, \& Lizhen Cui} 


\begin{abstract}
	Cross-modal alignment is an important multi-modal task, aiming to bridge the semantic gap between different modalities. The most reliable fundamention for achieving this objective lies in the semantic consistency between matched pairs. Conventional methods implicitly assume embeddings contain solely semantic information, ignoring the impact of non-semantic information during alignment, which inevitably leads to information bias or even loss. These non-semantic information primarily manifest as stylistic variations in the data, which we formally define as style information. An intuitive approach is to separate style from semantics, aligning only the semantic information. However, most existing methods distinguish them based on feature columns, which cannot represent the complex coupling relationship between semantic and style information. In this paper, we propose PICO, a novel framework for suppressing style interference during embedding interaction. Specifically, we quantify the probability of each feature column representing semantic information, and regard it as the weight during the embedding interaction. To ensure the reliability of the semantic probability, we propose a prototype iterative construction method. The key operation of this method is a performance feedback-based weighting function, and we have theoretically proven that the function can assign higher weight to prototypes that bring higher performance improvements. Extensive experiments on various benchmarks and model backbones demonstrate the superiority of PICO, outperforming state-of-the-art methods by 5.2\%-14.1\%. 
\end{abstract}




\maketitle
\section{Introduction}
\begin{figure}
	\centering
	\includegraphics[width=\linewidth]{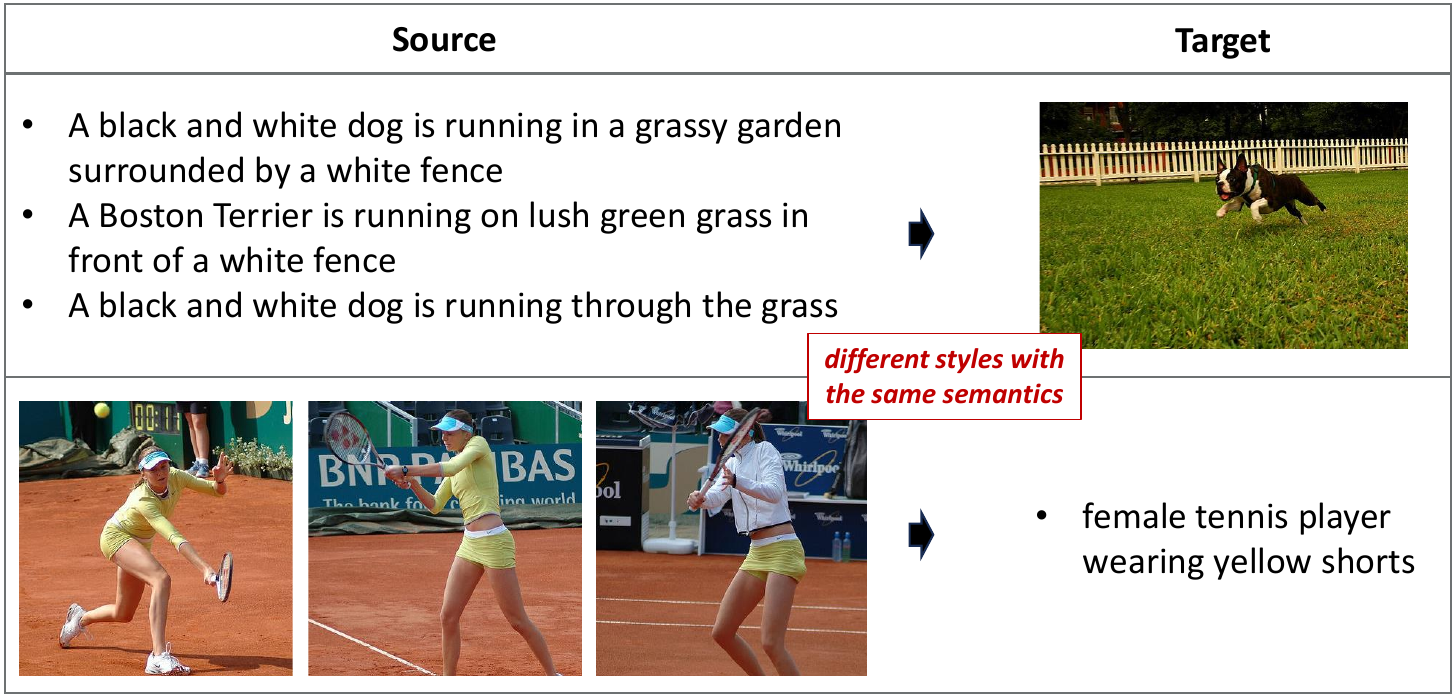}
	\caption{Images (or texts) with different expression styles can correspond to the same text (or image), indicating embeddings contain both semantic and non-semantic information.}
	\label{fig1}
\end{figure}
Cross-modal alignment is a crucial task in the field of multi-modal learning and serves as a foundational technique for tasks like image-text retrieval \cite{fu2024linguistic}, image captioning \cite{guo2019image,Visual2text}, and text-to-image generation \cite{li2019visual,liao2022text}. Its primary objective is to bridge the semantic gap between different modalities, such as vision and language. The key of this task lies in ensuring semantic consistency between image-text pairs, establishing correspondences between modalities.

Typically, cross-modal methods include two paradigms: coarse-grained and fine-grained methods. Coarse-grained alignment focuses on establishing global correspondences between modalities, matching entire images with their corresponding textual descriptions. Fine-grained alignment aims to align the regions in an image and words in the text. However, as shown in Fig.\ref{fig1}, images (or texts) with different expression styles may correspond to the same text (or image), indicating that the visual or textual embeddings contain not only semantic information but also non-semantic information. Such non-semantic information typically manifest as stylistic variations in texts or images, which we refer to as style information for clarity. The semantics of matching pairs can be aligned, whereas styles exhibit significant variations and cannot be precisely aligned. Conventional methods typically align the embeddings, ignoring the influence of style information, leading to information bias or even loss. Therefore, it is necessary to separate style from semantics and align only the semantics, to ensure the rationality and reliability of cross-modal alignment.

Most existing methods \cite{nlq} separate semantics and style by distinguishing feature columns of embeddings, assuming certain columns correspond to semantics while others represent style. These methods leverage intra-modal style consistency and cross-modal semantic consistency to impose constraints on feature columns for separation. In fact, semantics and style exhibit complex coupling relationships without clear evaluation criteria. Consequently, each feature column may simultaneously contains both semantic and style information, making simple column-based separation strategies unreliable.

To effectively suppress the interference of style information and improve reliability, we propose a reliable cross-modal alignment method based on \textbf{P}rototype \textbf{I}terative \textbf{CO}nstruction (\textbf{PICO}), as shown in Fig.\ref{framework}(c). By leveraging the semantic consistency information in image-text matched pairs, which is the most reliable supervisory knowledge for cross-modal alignment tasks, we quantify the probability of each feature column representing semantic information. Then, the semantic probability is regarded as the weight during the embedding interaction to achieving adaptive suppression of style-dominated feature columns. It should be clarified in cross-modal alignment tasks that the correlation scores of matched pairs need be optimized to higher values for establishing correspondences between modalities. The correlation scores are directly determined by the values of embedding interaction between feature columns. Positive interaction outputs enhance correlation scores, while negative values create suppression effects. Based on this, by statistically analyzing the sign distribution of interaction results, we can obtain pseudo-semantic probability for each feature column.

It is evident that these statistical results are highly susceptible to data partitioning or noise, exhibiting insufficient stability. We usually can evaluate the reliability of pseudo-semantic probability by constructing semantic prototypes and computing the divergence between feature columns and these prototypes. However, the particularity of our task lies in the extreme richness of semantics, which makes quantitative learning challenging. In contrast, the types of style demonstrate much greater consistency. Based on the fact that semantics and style probabilities are opposing events, we calculate semantic probability by first constructing style prototypes to obtain the style probability. During this process, to address the slow model convergence and prototype drift caused by excessive prototype variations across training epochs, we propose an iterative style prototype construction method. The core of iteration is designing appropriate weights of update strategy. We proposed a performance feedback-based weighting function, with theoretical guarantees that prototypes contributing more significantly to model improvement can be assigned higher update weights. Our contributions are summarized as follows:
\begin{enumerate}
	\item[$\bullet$] We propose a reliable cross-modal alignment method adaptively reduces the weights of feature columns dominated by style information during the embedding interaction.
	\item[$\bullet$] We introduce an iterative construction mechanism for style prototype, which explicitly represents style information and enhances the reliability of style prototypes.
	\item[$\bullet$] We propose a performance feedback-based dynamic weighting function for prototype updating, with theoretical guarantees it can adaptively assign higher weights to prototypes that contribute more to model performance improvement.
\end{enumerate}
\section{Related Work}
Current cross-modal alignment works can be broadly classified into coarse-grained and fine-grained methods \cite{fu2024linguistic}. \textbf{Coarse-grained methods} embed images and texts independently into a shared spacevia contrastive learning \cite{frome2013devise, li2019visual, wang2016learning}. Previous studies within this paradigm have frequently enhanced the joint embedding space by designing new losses \cite{faghrivse, chun2021probabilistic}, developing specialized architectures for backbones of different modalities \cite{wen2020learning, wu2019learning}, or learning better pooling strategies \cite{chen2021learning, li2022multi}. For instance, VSE++ \cite{faghrivse} introduced a triplet loss with hard negative mining, becoming a standard baseline for many following works. GPO \cite{chen2021learning} designs a new pooling operator that can learn from data. DIAS \cite{ma2024bridging} introduced spatial relationships between instances to improve alignment robustness. \textbf{Fine-grained methods} establish cross-modal interactions between image patches and text words, aggregating local matches into a global correlation score \cite{chen2020imram, diao2021similarity, lee2018stacked}. Unlike coarse-grained approaches, these methods explicitly model semantic correspondences between localized features. For example, SCAN \cite{lee2018stacked} is the first representative work that introduces cross-attention between the two modalities to find their alignments. NAAF \cite{zhang2022negative} adopted negative-aware learning to suppress mismatched pairs. CAAN \cite{zhang2020context} refines this concept by introducing an additional intra-modal interaction step following the cross-modal interaction. CHAN \cite{pan2023fine} addressed alignment noise through adaptive redundancy suppression, demonstrating improved robustness. 

However, these methods assumes the embeddings from different modalities interact only with the semantic information during embedding interaction. As mentioned earlier, embeddings contain both semantic and non-semantic information. Existing methods \cite{nlq} separate semantics and styles by decomposing feature columns, ignoring the complex coupling relationship between them. In this work, we focus on the calculation of the feature column's semantic probability, and improve the reliability of it through the prototype iterative construction and performance feedback-weight function.
\section{Methodology}
\begin{figure*}
	\centering
	\includegraphics[width=\linewidth]{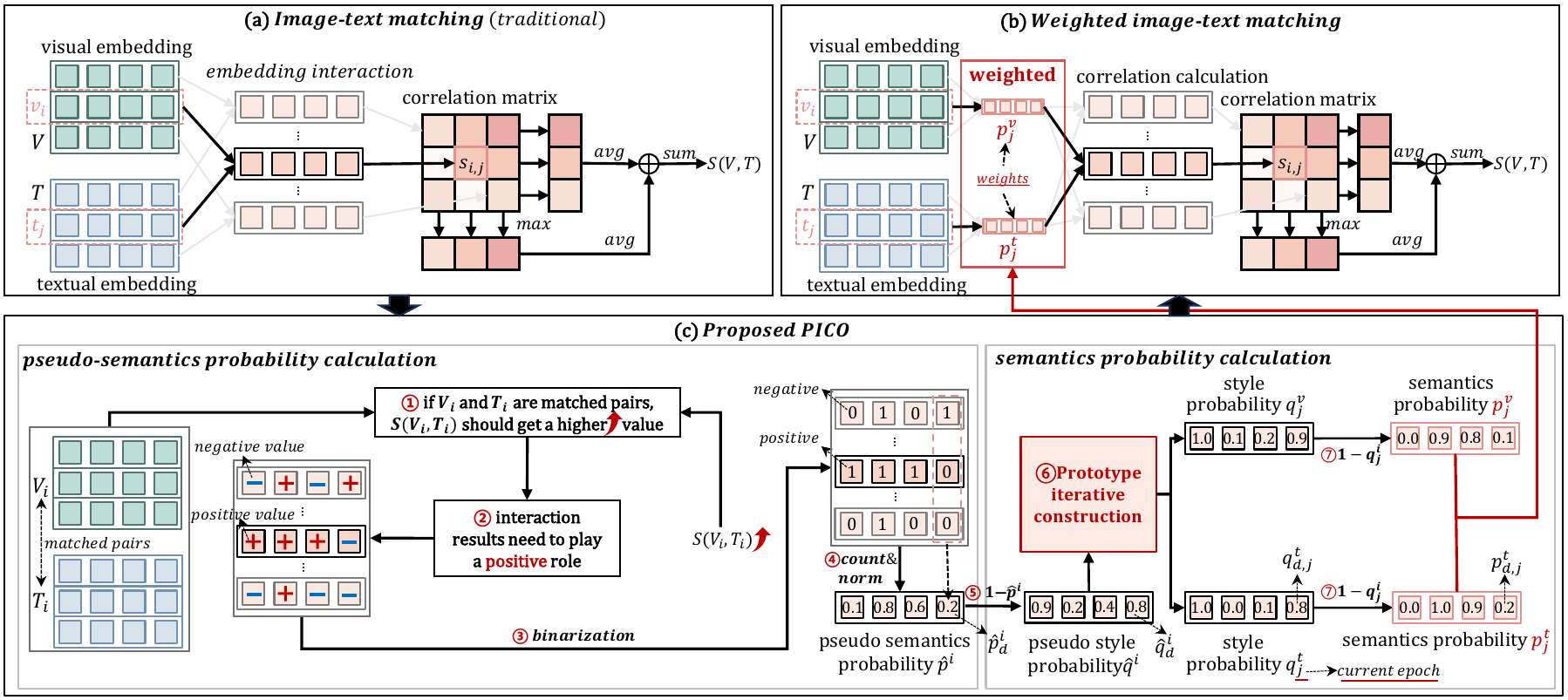}
	\caption{Overview of PICO. (a) Traditional fine-grained cross-modal alignment. (b) The weighted fine-grained cross-modal alignment, which weights feature columns during embedding interaction. (c) Semantic probability calculation of PICO. First, statistical analysis of feature column interactions yields pseudo-semantic and pseudo-style probabilities. Next, style prototypes are extracted and refined through iterative construction. Finally, comparing features with their prototypes provides style and semantic probabilities, where semantic probability weights suppress style-dominated features during embedding interaction.}
	\label{framework}
\end{figure*}
Considering effectiveness and interpretability, PICO adopts the fine-grained alignment method. Section \ref{3.1} introduces the framework of fine-grained alignment and highlights the differences between PICO and existing methods. Section \ref{3.2} and Section \ref{3.3} detail how to calculate pseudo-semantic probability and extract pseudo-style prototype. Section \ref{3.4} focuses on the prototype iterative construction, which is the core operation of PICO. The theoretical derivation of the prototype update strategy is also provided in section \ref{3.4}. Section \ref{3.5} details the calculation of semantic probability, and the objective function is described in section \ref{3.6}.
\subsection{The Framework of Fine-grained Alignment}\label{3.1}
We use the pure transformer architectures \cite{vaswani2017attention} to extract the visual and textual embeddings from image and text inputs, respectively. 

\textbf{Visual embeddings.} For an image $\textbf{V}$, we divide it into $n_v$ non-overlapping patches, and employ the vision transformer (ViT) \cite{alexey2020image, xu2022evo} to extract the visual embeddings of patches as $\textbf{V}=\{\textbf{v}_i|i\in [1,n_v],\textbf{v}_i\in \mathbb{R}^D\}$. $\textbf{v}_i$ means the visual embedding of $i$-th patch. $D$ is the number of feature columns.

\textbf{Textual embeddings.} Similarity, for a text (or sentence) $\textbf{T}$, we employ BERT \cite{devlin2018bert} to extract textual embeddings of words as $\textbf{T}=\{\textbf{t}_j|j\in [1,n_t],\textbf{t}_j\in \mathbb{R}^D\}$. $\textbf{t}_j$ means the textual embedding of $j$-th word. $n_t$ is the number of words in this text.

As shown in Fig.\ref{framework}(a), fine-grained alignment performs interactions between the visual and textual embeddings to obtain the correlation score $S(V,T)$. The maximum correspondence interaction is the commonly used approach \cite{fu2024linguistic}, formally as:
\begin{equation}
		S(V,T) = \sum_{i=1}^{n_v}\frac{\max(\{s_{i,j}|j\in[1,n_t]\})}{n_v}
		+\sum_{i=1}^{n_t}\frac{\max(\{s_{i,j}|i\in[1,n_v]\})}{n_t},
	\label{SVT}
\end{equation}
Here $\max(\cdot)$ means taking the maximum value. The first and secord terms are picking up the most aligned word for each patch and the most aligned patch for each word, and calculating the average of these corresponding correlation values to represent the correlation score $S(V,T)$ between image $\textbf{V}$ and text $\textbf{T}$. $s_{i,j}$ is the correlation value between patch $i$ and word $j$:
\begin{equation}
	s_{i,j} = \sum_{d=1}^{D} e_d = \sum_{d=1}^{D} v_{i,d}\cdot t_{j,d},
	\label{sij}
\end{equation}
Here $v_{i,d}$ and $t_{j,d}$ denote the $d$-th feature column`s value of patch $i$ and word $j$ respectively, and $e_d$ is the interaction result between $v_{i,d}$ and $t_{j,d}$. We define $\textbf{S}=\{s_{i,j}|i\in[1,n_v],j\in[1,n_t]\}$ as the correlation matrix. It can be realized that $\{e_d|d\in[1,D]\}$ directly determine $S(V,T)$, and the interaction results of different feature columns are treated equally. 

However, the information represented by feature columns includes semantic information and style information. Semantic information is alignable, whereas style information is not. Eq.\ref{sij} aligns all feature columns by default, which may lead to information bias or even loss. Thus, we improve Eq.\ref{sij} by weighting the interaction results, with the weights obtained by the semantic probability of each feature column. The semantic probability quantifies the probability of semantic information representation in the feature column.
\begin{equation}
	s_{i,j} = \sum_{d=1}^{D} e_d = \sum_{d=1}^{D} p^{v}_{d}v_{i,d}\cdot p^{t}_{d}t_{j,d},
	\label{oursij}
\end{equation} 
Here $p^{v}_{d}$ and $p^{t}_{d}$ are the semantic probability of $d$-th feature column in visual and textual embeddings, respectively. Define $\textbf{p}^{v}=\{p^{v}_{d}|d\in[1,D]\}$ and $p^{t}=\{p^{t}_{d}|d\in[1,D]\}$ are the semantic probability sets. Then, we can obtain the more reliable $S(V,T)$ via Eq.\ref{SVT}. Finally, The triplet loss is used as loss function to achieving cross-modality alignment, which can be expressed as:
\begin{equation}
	\mathcal{L}_{x}= [\alpha-S(V,T)+S(V,T^{-})]_{+}+[\alpha-S(V,T)+S(V^{-},T)]_{+},
	\label{trip}
\end{equation}
Here $\alpha$ is the margin parameter to control the degree of alignment, $[\cdot]_{+}=max(\cdot,0)$. $(V,T)$ is a positive image-text pair, and $(V,T^{-})$ and $(V^{-},T)$ are negative image-text pair in the batch.
\begin{figure*}
	\centering
	\includegraphics[width=\linewidth]{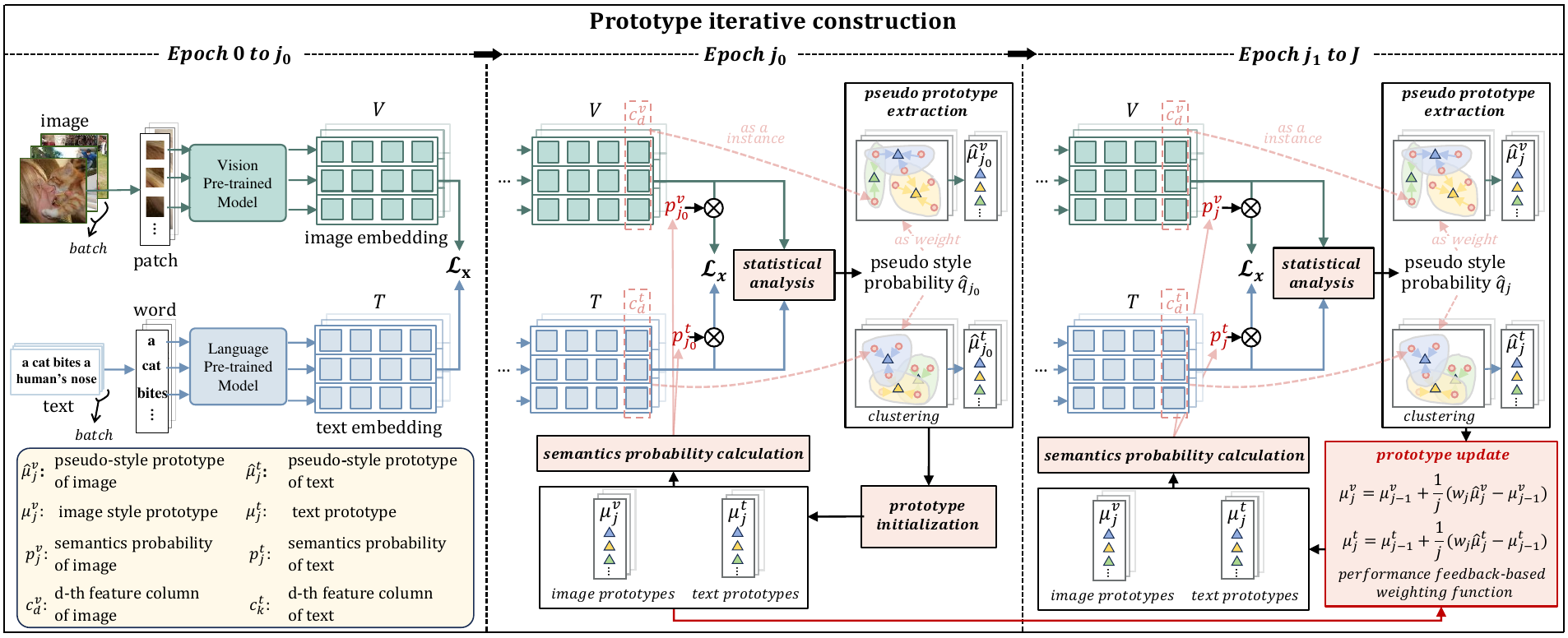}
	\caption{Prototype iterative construction. During epoch $0$ to $j_0$, visual-textual embedding alignment is initialized. At epoch $j_0$, pseudo-style prototypes are constructed by weighting feature columns with pseudo-style probabilities, serving as initial style prototypes. From epoch $j_1$ onward, these prototypes are iteratively updated via performance feedback-based weighting.}
	\label{pic}
\end{figure*}
\subsection{Pseudo-semantic Probability}\label{3.2}
Semantic and style information are entangled in the embeddings, with no intuitive criteria to clearly distinguish between them. Determining whether a feature column represents semantic or stylistic information is highly challenging, and simple column-based separation strategies are unreliable. So we calculate the semantic probability of each feature column based on reliable information within the data. 

The semantic consistency of image-text matched pairs is the most reliable supervisory knowledge for cross-modal alignment tasks, which means the correlation scores of matched pairs should be optimized to obtain higher values. As mentioned earlier, the correlation score $S(V,T)$ is directly related to the interaction results $e_d$. Positive $e_d$ can enhance $S(V,T)$, while negative $e_d$ creates suppression effects. To increase $S(V,T)$, $e_d$ should be as positive as possible. This implies that after initial learning, the feature columns with positive values in $e_d$ are more likely to represent semantic information. So, we construct the pseudo-semantic probability by statistically analyzing the positive and negative value distribution in all embeddings of each feature column.

For matched pairs $(V,T)$, the construction of pseudo-semantic probability can be expressed as:
\begin{equation}
	\hat{p}_d= \frac{1}{n_vn_t}\sum_{i=1}^{n_v}\sum_{j=1}^{n_t}B((e_d)_{i,j}),
	\label{bisum}
\end{equation}
Here $\hat{\textbf{p}}=\{\hat{p}_d|d\in[1,D]\}$ means the pseudo-semantic probability, and $\hat{p}_d$ is the pseudo-semantic probability of $d$-th feature column. $(e_d)_{i,j}$ means the $d$-th feature column's interaction result between patch $i$ and word $j$. $B(\cdot)$ is binary operation, set the value with positive sign to 1, otherwise it is 0.
\subsection{Pseudo-style Prototype Extraction}\label{3.3}
However, relying solely on statistical results can be unstable and unreliable due to data partitioning or noise interference. In general, we can evaluate the reliability of pseudo-semantic probability by constructing semantic prototypes and evaluating the difference between feature columns and semantic prototypes. The fundamental challenge lies in the inherent richness of semantics, making it difficult to learn quantitatively. Compared to semantics, style is more fixed in type. Therefore, we obtain the style probability by constructing style prototypes, and then calculate the semantic probability in reverse.

Based on the definitions of semantic probability and style probability, they are opposing events. Therefore, the pseudo-style probability is $\hat{q}_d = 1-\hat{p}_d$. Taking the construction of pseudo-style prototype for image modality as the instance, we define $\textbf{c}^v=\{\textbf{c}^v_d|d\in[1,D]\}$ as the feature column set of $V$, $\textbf{c}_d^v=\{v_{i,d}|i\in[1,n_v]\}$ denotes the $d$-th feature column. The pseudo-style probability of $\textbf{c}_d^v$ is $\hat{q}_d$. Then, We can implement Weighted K-means \cite{bezdek1984fcm} to construct pseudo-style prototypes with $\hat{q}_d$ as the weight and $\textbf{c}_d^v$ as the instance. To ensure the efficiency of training, we express the energy function $\mathcal{L}_{c}$ of clustering in the form of matrix operation:
\begin{equation}
	\mathcal{L}_{c}=Tr((\textbf{c}^v-\textbf{M}\hat{\mu}^v)^\top\hat{\textbf{q}}(\textbf{c}^v-\textbf{M}\hat{\mu}^v)),
	\label{kmeans}
\end{equation}
Here $\hat{\mu}^v=\{\hat{\mu}^v_k|k\in[1,K]\}$ is the cluster center matrix of image modality, $\hat{\mu}^v_k\in \mathbb{R}^{D}$ is the center of cluster $k$. $K$ is the number of clusters. $\hat{\textbf{q}}=diag(\hat{\textbf{q}}_1,\hat{\textbf{q}}_2,\dots,\hat{\textbf{q}}_D)\in\mathbb{R}^{D\times D}$ is the weight matrix. $Tr(\cdot)$ is the trace of the matrix. $\textbf{M}\in \mathbb{R}^{D\times K}$ is the indicator matrix to indicate the membership of instances, which is a learnable binary matrix. $M_{d,k}=1$ means instance $\textbf{c}_d^v$ belong to cluster $k$. $\textbf{c}^v-\textbf{M}\hat{\mu}^v$ represents the difference matrix between all instances and their cluster centers. $(\textbf{c}^v-\textbf{M}\hat{\mu}^v)^\top\hat{\textbf{q}}(\textbf{c}^v-\textbf{M}\hat{\mu}^v)$ is the weighted covariance matrix. The update function of cluster center is:
\begin{equation}
	\hat{\mu}^v=\frac{\textbf{M}^\top\hat{\textbf{q}}\textbf{c}^v}{\textbf{M}^\top\hat{\textbf{q}}\textbf{M}},
	\label{center}
\end{equation}
The cluster center represents the typical characteristics of the whole cluster, so we can take $\hat{\mu}^v$ as the pseudo-style prototypes of image modality. Following the similar approach, we can also obtain the pseudo-style prototypes of text modality $\hat{\mu}^t$.

\subsection{Prototype Iterative Construction}\label{3.4}
Clustering-based prototypes can capture the typical characteristics of instances, but there are also limitations. First, since clustering is performed anew in each epoch, the resulting prototypes may vary greatly across epochs. This inconsistency can lead to oscillations during training, significantly slowing down model convergence. Second, the prototypes obtained in each epoch are sensitive to parameter initialization and noisy instances, and outliers can induce prototype drift, thereby reducing their representational validity.

We propose a novel prototype iterative construction method to avoid the problems in conventional methods, as shown in Fig.\ref{pic}. The method weights and aggregates the pseudo-style prototypes of all epochs into the style prototypes, and performs a iterative update strategy to gradually improve the effectiveness of style prototypes, ensuring robust and stable representation learning. Specifically, we first train the model for $j_0$ epochs based on Eq.\ref{trip} to achieve preliminary alignment between visual and textual embeddings. This ensures the validity of pseudo-style prototypes constructed in clustering operations. At the epoch $j_0$, we compute $\hat{\mu}^v_{j_0}$ via Eq.\ref{center} and serves as the initial value for the style prototype. $\hat{\mu}^v_{j_0}$ means the pseudo-style prototype of epoch $j_0$. From epoch $j_1$ until the final training epoch $J$, we employ the following update strategy: First, compute the current epoch's pseudo-style prototype $\hat{\mu}^v_{j}$ by Eq.\ref{center}. Then, update the style prototype according to the update function:
\begin{equation}
	\mu_j^v = \mu_{j-1}^v + \frac{1}{j}(w_{j}\hat{\mu}^v_{j}-\mu_{j-1}^v), \ \
	j\in[j_0,J],
	\label{update}
\end{equation}
Here $j$ is the number of current epoch. $\mu_j^v=\{\mu^v_{j,k}|k\in[1,K]\}$ represents the pseudo-style prototype of epoch $j$, and $\mu^v_{j,k}$ is the $k$-th pseudo-style prototype. For assigning higher weight to prototypes that bring higher performance improvements, we proposed the performance feedback-based weighting function:
\begin{equation}
	w_{j} = 1+\frac{1}{\bar{rSum}_{j_0:j-1}}(rSum_{j-1}-rSum_{j-2}),
	\label{wt}
\end{equation}
$rSum_{t-1}$ means the sum of Recall at K (R@K) at epoch $j-1$, which is the main evaluation metric of cross-modality alignment. $\bar{rSum}_{j_0:j-1}$ is average of the rSum from epoch $j_0$ to $j-1$. The value of $rSum_{t-1}$ is a quantitative indicator for evaluating the effectiveness of the style prototype $\mu_{j-1}^v$ to some extent, while $w_{t}$ can measure the performance improvement achieved through the style prototype update from $\mu_{j-2}^v$ to $\mu_{j-1}^v$. The \textbf{proof} proceeds as follows: 
\begin{equation}
	\begin{aligned}
		\mu_j^v &= \mu_{j-1}^v + \frac{1}{j}(w_{j}\hat{\mu}^v_{j}-\mu_{j-1}^v)=\frac{j-1}{j}\mu_{j-1}^v+\frac{w_{j}}{j}\hat{\mu}^v_{j}\\
		&=\frac{j-2}{j}\mu_{j-2}^v+\frac{w_{j-1}}{j-1}\hat{\mu}^v_{j-1}+\frac{w_{j}}{j}\hat{\mu}^v_{j}\\
		&=\frac{1}{j}(w_{j_0}\hat{\mu}_{j_0}^v+w_{j_0+1}\hat{\mu}_{j_0+1}^v+\cdots+w_{j}\hat{\mu}^v_{j}).
		\label{proof}
	\end{aligned}
\end{equation}
So the weight of the pseudo-style prototype is directly determined by the performance feedback. Pseudo-style prototypes with well results perform greater influence on the current epoch's style prototype, ensuring the effectiveness and robust of the update strategy. Eq.\ref{wt} can ensure equitable contribution across epochs, while adaptively increasing the weight of pseudo-style prototypes corresponding to epochs with significant rSum improvement. 
\subsection{Semantic Probability}\label{3.5}
After constructing the style prototypes, we can calculate the distance between a given instance and its assigned style prototype. The distance quantifies the style probability of that instance. Specifically, a smaller distance indicates a higher probability that the instance aligns with the style represented by the style prototype. 

Assuming that $\textbf{c}^v_d$ is most similar to $k$-th style prototype $\mu^v_{k}$, the style probability can be expressed as following. For the convenience of introduction, we have omitted the number of epoch $j$:
\begin{equation}
	q_d^v=sigmoid(\frac{1}{\varepsilon}||\textbf{c}^v_d-\mu_k^v||_2^2),
	\label{style}
\end{equation}
Here $\textbf{q}^v=\{q_d^v|d\in[1,D]\}$ is the style probability of $\textbf{c}^v$, and $q_d^v$ is the style probability of $\textbf{c}^v_d$. $\varepsilon$ is a adjustment parameter. According to the fact that semantic probability and style probability are opposing events, we can obtain the semantic probability of image modality $p^v_d = 1-q^v_d$, and $\textbf{p}^v=\{p_d^v|d\in[1,D]\}$ Following a similar approach, we can also obtain the semantic probability of text modality $\textbf{p}^t$, the style probability of text modality $\textbf{q}^t$.
\subsection{Objective Function}\label{3.6}
We combine the triplet loss and energy function of clustering as the loss function $\mathcal{L}$ of PICO:
\begin{equation}
	\mathcal{L}=\mathcal{L}_{x}+\omega_c\mathcal{L}_{c},
	\label{floss}
\end{equation}
Here $\omega_c$ is the hyper-parameter to control the cluster compactness during the clustering operation. Note that $\omega_c$ remains 0 for the first $j_0$ epochs to ensure the implementation of prototype iterative construction. We use the distance weighted sampling \cite{wu2017sampling} for hard negative mining to ensure learning efficiency. 
\section{Experiments}
\subsection{Experimental Setup}
\begin{table*}
	\centering
	\caption{The comparisons of image-text retrieval performances with state-of-the-art methods on Flickr30K and MS-COCO. We list the backbones, image resolution, and the number of patches (e.g., The `ViT-224 + BERT, $14\times14$ patches' means the base-version of ViT\cite{alexey2020image} with $224\times224$ image resolution input, getting $14\times14$ patches for one image, and the base-version of BERT\cite{devlin2018bert} for text words). The best results are marked \textbf{bold}. `$*$' denotes the coarse-grained method.}
	\resizebox{\textwidth}{!}{
		\begin{tabular}{l|ccccccc|ccccccc|ccccccc}
			\hline
			\multirow{3}{*}{Methods}&\multicolumn{7}{c|}{Flickr30K 1K}&\multicolumn{7}{c|}{MS-COCO 1K}&\multicolumn{7}{c}{MS-COCO 5K}\\
			&\multicolumn{3}{c}{Image-to-Text}&\multicolumn{3}{c}{Text-to-Image}&rSum&\multicolumn{3}{c}{Image-to-Text}&\multicolumn{3}{c}{Text-to-Image}&rSum&\multicolumn{3}{c}{Image-to-Text}&\multicolumn{3}{c}{Text-to-Image}&rSum\\
			&R@1&R@5&R@10&R@1&R@5&R@10&&R@1&R@5&R@10&R@1&R@5&R@10&&R@1&R@5&R@10&R@1&R@5&R@10&\\
			\hline
			\multicolumn{22}{l}{\textbf{\textit{FasterR-CNN \cite{ren2016faster} + BERT}, 36 pre-computed regions}}\\
			DIAS\cite{ma2024bridging}&83.8&96.6&98.3&64.5&88.0&93.3&524.5&83.4&97.1&99.1&67.6&92.4&96.6&536.2&64.4&88.9&94.1&47.2&76.5&85.2&456.3\\
			HREM$^*$\cite{fu2023learning}&83.3&96.0&98.1&63.5&87.1&92.4&520.4&81.1&96.6&98.9&66.1&91.6&96.5&530.7&62.3&87.6&93.4&43.9&73.6&83.3&444.1\\
			CHAN\cite{pan2023fine}&80.6&96.1&97.8&63.9&87.5&92.6&518.5&81.4&96.9&98.9&66.5&92.1&96.7&532.6&59.8&87.2&93.3&44.9&74.5&84.2&443.9\\
			\hline
			\multicolumn{22}{l}{\textbf{\textit{ViT-224 + BERT}, 14$\times$14 patches}}\\
			VSE++$^*$\cite{faghrivse}&71.8&92.8&96.5&59.4&84.7&90.9&496.1&75.0&94.6&98.0&62.7&89.4&94.9&514.6&52.4&80.3&88.8&40.6&70.4&81.1&413.4\\
			SCAN\cite{lee2018stacked}&69.5&90.9&95.6&56.4&83.1&90.0&485.6&76.0&95.4&98.1&64.5&90.8&95.8&520.6&53.9&81.8&90.0&42.9&72.3&82.5&423.5\\
			SGR\cite{diao2021similarity}&69.7&90.8&95.2&59.1&84.1&89.9&488.7&77.2&95.0&98.0&65.1&90.7&95.8&521.8&54.9&82.8&90.5&42.8&72.2&82.5&425.8\\
			CHAN\cite{pan2023fine}&69.2&91.8&95.0&58.4&84.9&90.6&489.9&77.1&95.1&98.1&65.0&91.0&96.0&522.2&56.3&83.2&90.1&43.0&72.6&82.8&428.0\\
			LAPS \cite{fu2024linguistic}&74.0&93.4&97.4&62.5&87.3&92.7&507.3&78.7&95.5&98.3&66.2&91.3&96.2&526.3&\textbf{57.5}&84.0&90.8&44.5&74.0&83.6&434.4\\
			\textbf{PICO}&\textbf{74.5}&\textbf{94.0}&\textbf{98.2}&\textbf{63.0}&\textbf{88.5}&\textbf{93.1}&\textbf{511.3}&\textbf{78.8}&\textbf{95.9}&\textbf{98.8}&\textbf{66.3}&\textbf{91.6}&\textbf{96.5}&\textbf{527.9}&\textbf{57.5}&\textbf{84.1}&\textbf{91.2}&\textbf{44.9}&\textbf{74.3}&\textbf{83.8}&\textbf{435.8}\\
			\hline
			\multicolumn{22}{l}{\textbf{\textit{ViT-384 + BERT}, 24$\times$24 patches}}\\
			VSE++$^*$\cite{faghrivse}&77.1&95.7&97.5&65.8&90.2&94.3&520.5&77.0&95.7&98.4&64.6&91.1&96.2&523.0&54.9&82.8&90.4&42.4&72.4&82.8&425.8\\
			SCAN\cite{lee2018stacked}&75.4&94.4&96.9&63.6&88.6&93.5&512.5&76.1&95.5&98.5&65.1&91.6&96.3&523.1&53.3&81.8&90.0&42.6&72.6&82.9&423.1\\
			SGR\cite{diao2021similarity}&76.9&94.9&98.1&64.2&88.4&93.3&515.8&75.8&95.7&98.6&65.6&92.0&96.5&524.2&53.3&81.0&89.6&42.9&73.1&83.7&423.6\\
			CHAN\cite{pan2023fine}&75.4&94.5&97.6&63.2&88.6&93.1&512.4&78.1&95.8&98.6&66.1&92.1&96.6&527.3&55.6&83.8&91.2&43.4&73.6&83.5&431.1\\
			LAPS \cite{fu2024linguistic}&79.0&96.0&98.1&67.3&90.5&94.5&525.4&78.7&96.3&98.9&68.0&92.4&96.8&531.0&57.4&84.9&92.5&46.4&75.8&85.2&442.2\\
			\textbf{PICO}&\textbf{79.1}&\textbf{96.3}&\textbf{98.2}&\textbf{67.5}&\textbf{90.9}&\textbf{94.7}&\textbf{526.7}&\textbf{78.9}&\textbf{96.5}&\textbf{98.9}&\textbf{68.2}&\textbf{92.7}&\textbf{96.9}&\textbf{532.1}&\textbf{57.7}&\textbf{85.1}&\textbf{92.9}&\textbf{46.7}&\textbf{76.0}&\textbf{85.6}&\textbf{444.0}\\
			\hline
			\multicolumn{22}{l}{\textbf{\textit{Swin-224 + BERT}, 7$\times$7 patches}}\\
			VSE++$^*$\cite{faghrivse}&82.5&96.5&98.9&70.0&91.4&95.1&534.4&83.3&97.5&99.3&71.0&93.0&96.7&540.9&64.0&88.2&94.2&49.9&78.0&86.6&460.9\\
			SCAN\cite{lee2018stacked}&79.0&95.9&98.2&67.7&90.6&94.9&526.3&80.9&97.0&99.1&69.7&93.1&97.1&536.9&60.7&86.6&93.2&48.1&77.1&86.1&451.8\\
			SGR\cite{diao2021similarity}&80.4&97.0&98.7&66.9&90.2&94.5&527.6&81.2&97.1&99.1&69.9&93.2&97.2&537.7&61.0&86.7&93.2&48.6&77.2&86.3&453.1\\
			CHAN\cite{pan2023fine}&81.4&97.0&98.6&68.5&90.6&94.5&530.6&81.6&97.2&99.3&70.6&93.7&97.6&539.8&64.1&87.9&93.5&49.1&77.3&86.1&458.0\\
			LAPS \cite{fu2024linguistic}&82.4&97.4&99.5&70.0&91.7&95.4&536.3&84.0&97.6&99.3&\textbf{72.1}&93.7&97.3&544.1&64.5&89.2&94.4&51.6&78.9&87.2&465.8\\
			\textbf{PICO}&\textbf{82.9}&\textbf{97.9}&\textbf{99.6}&\textbf{70.3}&\textbf{92.2}&\textbf{95.6}&\textbf{538.5}&\textbf{84.2}&\textbf{97.9}&\textbf{99.5}&\textbf{72.1}&\textbf{93.8}&\textbf{97.4}&\textbf{544.9}&\textbf{64.6}&\textbf{89.7}&\textbf{94.8}&\textbf{51.7}&\textbf{79.2}&\textbf{87.5}&\textbf{467.5}\\
			\hline
			\multicolumn{22}{l}{\textbf{\textit{Swin-384 + BERT}, 12$\times$12 patches}}\\
			VSE++$^*$\cite{faghrivse}&83.3&97.5&99.2&71.1&93.2&96.2&540.6&82.9&97.7&99.4&71.3&93.5&97.3&542.1&63.0&88.5&94.3&50.1&78.9&87.4&462.2\\
			SCAN\cite{lee2018stacked}&81.9&96.9&98.9&70.0&92.7&95.8&536.1&81.6&96.8&99.1&69.1&92.7&96.7&536.1&61.1&87.3&93.3&47.8&76.9&85.9&452.4\\
			SGR\cite{diao2021similarity}&80.7&96.8&99.0&69.9&91.7&95.3&533.4&81.9&96.7&99.1&69.3&92.8&96.7&536.6&62.8&87.0&92.9&48.1&77.0&86.0&453.8\\
			CHAN\cite{pan2023fine}&81.2&96.7&98.8&70.3&92.2&95.9&535.0&83.1&97.3&99.2&70.4&93.1&97.1&540.2&63.4&88.4&94.1&49.2&77.9&86.6&459.5\\
			LAPS \cite{fu2024linguistic}&85.1&97.7&99.2&74.0&93.0&96.3&545.3&84.1&97.4&99.2&72.1&93.9&97.4&544.1&67.1&88.6&94.3&53.0&79.5&87.6&470.1\\
			\textbf{PICO}&\textbf{85.8}&\textbf{98.1}&\textbf{99.4}&\textbf{74.5}&\textbf{93.5}&\textbf{96.9}&\textbf{548.2}&\textbf{84.4}&\textbf{97.8}&\textbf{99.5}&\textbf{72.5}&\textbf{94.3}&\textbf{97.9}&\textbf{546.4}&\textbf{67.4}&\textbf{89.0}&\textbf{94.5}&\textbf{53.1}&\textbf{79.8}&\textbf{88.0}&\textbf{471.8}\\
			\hline
	\end{tabular}}
	\label{sota}
\end{table*}
\begin{table}
	\centering
	\caption{The comparisons of image-text retrieval performances with vision-language pre-training (VLP) Models. `$\#$' is the zero-shot learning. `\textbf{\textit{Large}}' means the large-version. }
	\resizebox{0.48\textwidth}{!}{\begin{tabular}{l|cccc|cccc}
			\hline
			\multirow{3}{*}{Methods}& \multicolumn{4}{c|}{Flickr30K 1K}  & \multicolumn{4}{c}{MS-COCO 5K}\\
			&\multicolumn{2}{c}{Image-to-Text}&\multicolumn{2}{c|}{Text-to-Image}&\multicolumn{2}{c}{Image-to-Text}&\multicolumn{2}{c}{Text-to-Image}\\
			&R@1&R@5&R@1&R@5&R@1&R@5&R@1&R@5\\
			\hline
			VILT \cite{chen2020uniter}&83.5&96.7&64.4&88.7&61.5&86.3&42.7&72.9\\
			SOHO \cite{kim2021vilt}&86.5&98.1&72.5&92.7&66.4&88.2&50.6&78.0\\
			ALBEF \cite{huang2021seeing}&95.9&99.8&85.6&97.5&77.6&94.3&60.7&84.3\\
			BLIP \cite{li2022blip}&96.6&99.8&87.2&97.5&80.6&95.2&63.1&85.3\\
			\hline
			\multicolumn{9}{l}{\textbf{\textit{CLIP-ViT-224 + CLIP-BERT}, 14$\times$14 patches}}\\
			CLIP$^{\#*}$ \cite{radford2021learning}&81.4&96.2&61.1&85.4&52.3&76.2&33.3&58.2\\
			VSE++$^*$ \cite{faghrivse}&92.2&99.1&80.5&95.6&66.8&88.2&53.6&79.7\\
			SCAN \cite{lee2018stacked}&88.2&98.1&75.3&93.1&65.4&88.0&50.7&77.6\\
			LAPS \cite{fu2024linguistic}&92.9&99.3&80.6&95.5&69.8&90.4&54.3&80.0\\
			\textbf{PICO} &\textbf{93.2}&\textbf{99.4}&\textbf{81.3}&\textbf{96.2}&\textbf{70.4}&\textbf{90.8}&\textbf{54.8}&\textbf{80.6}\\
			\hline
			\multicolumn{9}{l}{\textbf{\textit{CLIP-ViT-Large-224 + CLIP-BERT-Large}, 16$\times$16 patches}}\\
			CLIP$^{\#*}$ \cite{radford2021learning}&85.0&97.7&64.3&87.0&55.9&79.1&35.9&60.9\\
			VSE++$^*$ \cite{faghrivse}&94.0&99.5&83.4&96.4&68.5&89.4&56.7&81.9\\
			SCAN \cite{lee2018stacked}&90.0&98.5&81.0&95.9&68.0&90.4&53.2&80.7\\
			LAPS \cite{fu2024linguistic}&94.6&\textbf{99.9}&84.9&97.3&72.9&91.7&57.1&81.3\\
			\textbf{PICO}&\textbf{95.0}&\textbf{99.9}&\textbf{85.4}&\textbf{97.9}&\textbf{73.4}&\textbf{92.2}&\textbf{57.3}&\textbf{81.9}\\
			\hline
	\end{tabular}}
	\label{sota2}
\end{table}
\textbf{Datasets and Metrics.} Following the previous works \cite{fu2024linguistic, ma2024bridging}, we evaluate PICO mainly on the Flickr30K \cite{young2014image} and MS-COCO \cite{lin2014microsoft} datasets. Flickr30k contains 29,000, 1,000, and 1,000 images for training, testing, and validation. MS-COCO contains 82,738, 5,000, and 5,000 images for training, testing, and validation. Each image is associated with 5 texts. The results on MS-COCO are reported on averaging over 5-folds of 1,000 test images and on the full 5,000 test images. The Recall at K (R@K) and rSum are adopted as the evaluation metrics. R@K means the percentage of ground truth in the retrieved top-K lists, and K=1,5,10. rSum reflects the overall performance, which is the sum of multiple R@K in both image-to-text and text-to-image alignments.

\textbf{Implementation details.} We use the Vision Transformer (ViT) \cite{alexey2020image} and Swin Transformer (Swin) \cite{liu2021swin} as backbones to extract visual embeddings, and use BERT \cite{devlin2018bert} to extract textual embeddings. The experimental setting are based on the backbones's base version. A patch is $16\times16$ pixels for ViT, and is $32\times32$ pixels for Swin. The image resolutions are $224\times224$ and $384\times384$. So there are $14\times14$ and $24\times24$ patches for ViT, and $7\times7$ and $12\times12$ patches for Swin. An additional linear layer is introduced on the top of these backbones to unify feature size $D$ as 512. The whole framework is trained for 30 epochs on a NVIDIA L40 GPU. AdamW optimizer \cite{loshchilov2017decoupled, zhang2018improved} is adopted with learning rate of $2e^{-4}$. The batch size is 64.
\begin{figure*}
	\centering
	\includegraphics[width=\linewidth]{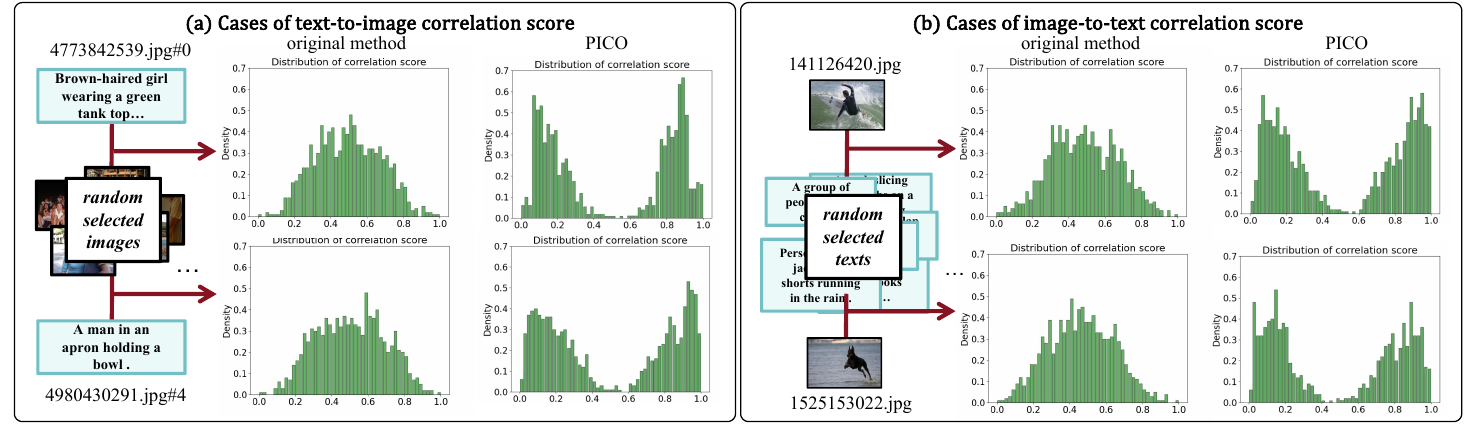}
	\caption{The visualization of correlation score'1 distribution with / without PICO's weighting process during embedding interaction. After weighting, correlation scores are more closer to both ends (0 or 1), simplifying match assessment.}
	\label{viscas}
\end{figure*}
\begin{table}
	\centering
	\caption{Ablation studies of PICO. $CR$ denotes the change rate.}
	\resizebox{0.48\textwidth}{!}{\begin{tabular}{l|cccc|cccc|c}
			\hline
			\multirow{3}{*}{Methods}& \multicolumn{4}{c|}{Flickr30K 1K}  & \multicolumn{4}{c|}{MS-COCO 5K}&\\
			&\multicolumn{2}{c}{Image-to-Text}&\multicolumn{2}{c|}{Text-to-Image}&\multicolumn{2}{c}{Image-to-Text}&\multicolumn{2}{c|}{Text-to-Image}&$CR$\\
			&R@1&R@5&R@1&R@5&R@1&R@5&R@1&R@5&\\
			\hline
			\multicolumn{10}{l}{\textbf{\textit{ViT-224 + BERT}, 14$\times$14 patches}}\\
			w/o $Wei$&68.9&90.8&55.9&80.7&52.2&80.3&40.3&69.9&\textcolor{blue}{-7.2\%}\\
			w/o $Pro$&70.2&90.9&59.3&83.1&54.6&81.2&42.5&71.2&\textcolor{blue}{-4.8\%}\\
			w/o $Ite$&71.9&91.2&60.8&83.7&55.0&82.2&42.9&72.1&\textcolor{blue}{-3.6\%}\\
			w/o $Fed$&73.8&93.2&61.9&87.4&55.9&83.5&43.8&72.9&\textcolor{blue}{-0.9\%}\\
			\textbf{PICO}&\textbf{74.5}&\textbf{94.0}&\textbf{63.0}&\textbf{88.5}&\textbf{57.5}&\textbf{84.1}&\textbf{44.9}&\textbf{74.3}&--\\
			\hline
			\multicolumn{10}{l}{\textbf{\textit{Swin-224 + BERT}, 7$\times$7 patches}}\\
			w/o $Wei$&78.8&92.1&65.2&89.5&60.9&85.7&47.5&76.5&\textcolor{blue}{-9.4\%}\\
			w/o $Pro$&80.3&95.2&68.0&90.2&62.3&86.2&48.3&77.1&\textcolor{blue}{-3.3\%}\\
			w/o $Ite$&81.4&96.0&69.1&90.7&63.5&87.9&49.6&77.8&\textcolor{blue}{-2.0\%}\\
			w/o $Fed$&82.3&96.8&69.5&91.2&64.0&88.6&51.1&78.6&\textcolor{blue}{-1.0\%}\\
			\textbf{PICO}&\textbf{82.9}&\textbf{97.9}&\textbf{70.3}&\textbf{92.2}&\textbf{64.6}&\textbf{89.7}&\textbf{51.7}&\textbf{79.2}&--\\
			\bottomrule
	\end{tabular}}
	\label{ab}
\end{table}
\begin{table}
	\centering
	\caption{The application effect of distribution sampling method to other models with backbone `\textbf{\textit{ViT-224}}'.}
	\resizebox{0.48\textwidth}{!}{\begin{tabular}{l|cccc|cccc|c}
			\hline
			\multirow{3}{*}{Methods}& \multicolumn{4}{c|}{Flickr30K 1K}  & \multicolumn{4}{c|}{MS-COCO 5K}&\\
			&\multicolumn{2}{c}{Image-to-Text}&\multicolumn{2}{c|}{Text-to-Image}&\multicolumn{2}{c}{Image-to-Text}&\multicolumn{2}{c|}{Text-to-Image}&$CR$\\
			&R@1&R@5&R@1&R@5&R@1&R@5&R@1&R@5&\\
			\hline
			VSE++\cite{faghrivse}&71.8&92.8&59.4&84.7&52.4&80.3&40.6&70.4&\multirow{2}{*}{\textcolor{red}{+3.6\%}}\\
			\textbf{+PICO}&\textbf{73.1}&\textbf{93.2}&\textbf{64.9}&\textbf{86.7}&\textbf{55.5}&\textbf{83.1}&\textbf{43.0}&\textbf{72.9}&\\
			\hline
			SCAN\cite{lee2018stacked}&69.5&90.9&56.4&83.1&53.9&81.8&42.9&72.3&\multirow{2}{*}{\textcolor{red}{+2.0\%}}\\
			\textbf{+PICO}&\textbf{72.4}&\textbf{91.5}&\textbf{57.9}&\textbf{84.8}&\textbf{54.7}&\textbf{83.2}&\textbf{43.9}&\textbf{73.2}&\\
			\hline
			SGR\cite{diao2021similarity}&69.7&90.8&59.1&84.1&54.9&82.8&42.8&72.2&\multirow{2}{*}{\textcolor{red}{+1.8\%}}\\
			\textbf{+PICO}&\textbf{72.5}&\textbf{91.7}&\textbf{60.2}&\textbf{86.0}&\textbf{55.1}&\textbf{83.6}&\textbf{43.9}&\textbf{73.3}&\\
			\hline
			CHAN\cite{pan2023fine}&69.2&91.8&58.4&84.9&56.3&83.2&43.0&72.6&\multirow{2}{*}{\textcolor{red}{+1.6\%}}\\
			\textbf{+PICO}&\textbf{71.8}&\textbf{92.8}&\textbf{59.1}&\textbf{86.7}&\textbf{56.9}&\textbf{83.7}&\textbf{44.1}&\textbf{73.3}&\\
			\hline
			LAPS \cite{fu2024linguistic}&74.0&93.4&62.5&87.3&57.5&84.0&44.5&74.0&\multirow{2}{*}{\textcolor{red}{+0.9\%}}\\
			\textbf{+PICO}&\textbf{74.9}&\textbf{94.0}&\textbf{62.9}&\textbf{88.5}&\textbf{57.9}&\textbf{84.3}&\textbf{45.1}&\textbf{74.7}&\\
			\hline
	\end{tabular}}
	\label{cross}
\end{table}
\subsection{Comparison with State-of-the-art Methods}
To show the performance superiority of PICO, we compare it with state-of-the-art (SOTA) methods on the two datasets. The results of DIAS \cite{ma2024bridging}, HREM \cite{fu2023learning} and CHAN \cite{pan2023fine} are cited directly from their original publications, while all other methods are implemented using their official source codes to generate comparable results. As shown in Tab.\ref{sota}, we persent quantitative results on Flickr30K and MS-COCO datasets. Our model outperformers SOTA methods with impressive margins on the R@K and rSum, and achieves consistent superiority on different backbones. Notably, enhanced performance is observed when employing more sophisticated transformer-based backbones, as measured by both the architectural depth and the number of input patches.

To further demonstrate the performance, we extend our architecture to the classic Vision-Language Pre-training (VLP) model CLIP \cite{radford2021learning} and the current SOTA VLP models \cite{chen2020uniter,kim2021vilt,li2022blip}, as shown in Tab.\ref{sota2}. The experimental results reveal that current fine-grained methods, despite leveraging VLP backbones, still struggle to achieve satisfactory results. In contrast, PICO achieves significant improvements and demonstrating competitive performance compared to the mainstream VLP models.
\begin{figure}
	\centering
	\includegraphics[width=\linewidth]{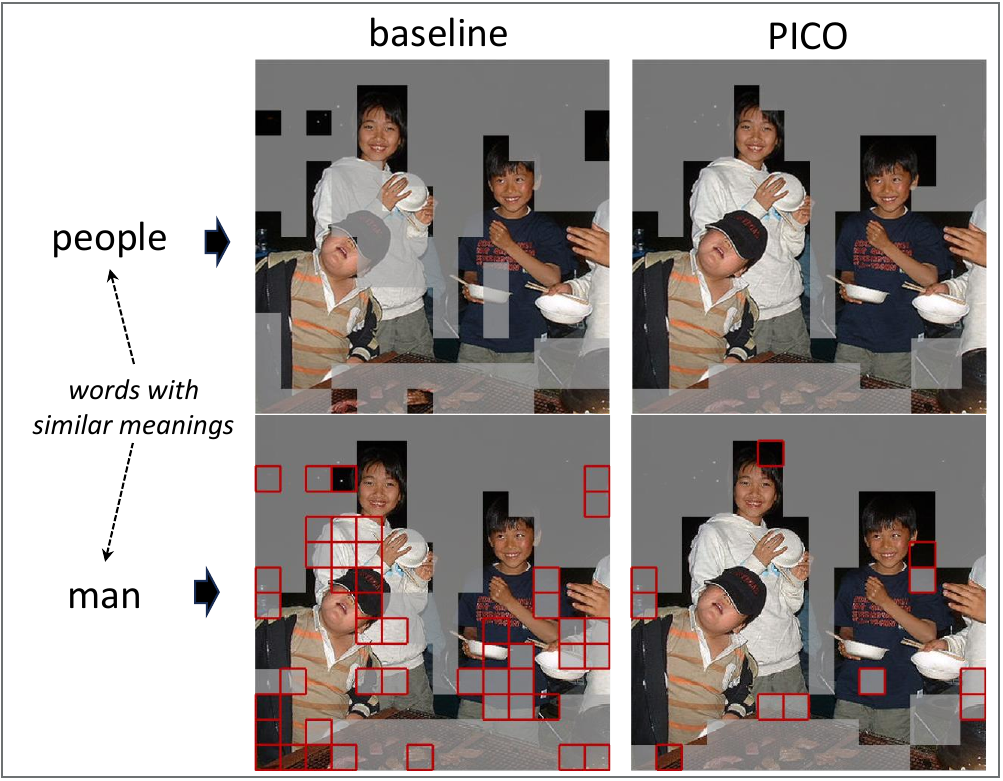}
	\caption{The visualization of patches corresponding to words with similar meanings. The red boxes indicate the differences between patches selected by the two methods.}
	\label{vis}
\end{figure}
\subsection{Ablation Study and Discussion}
\begin{figure*}
	\centering
	\includegraphics[width=0.33\linewidth]{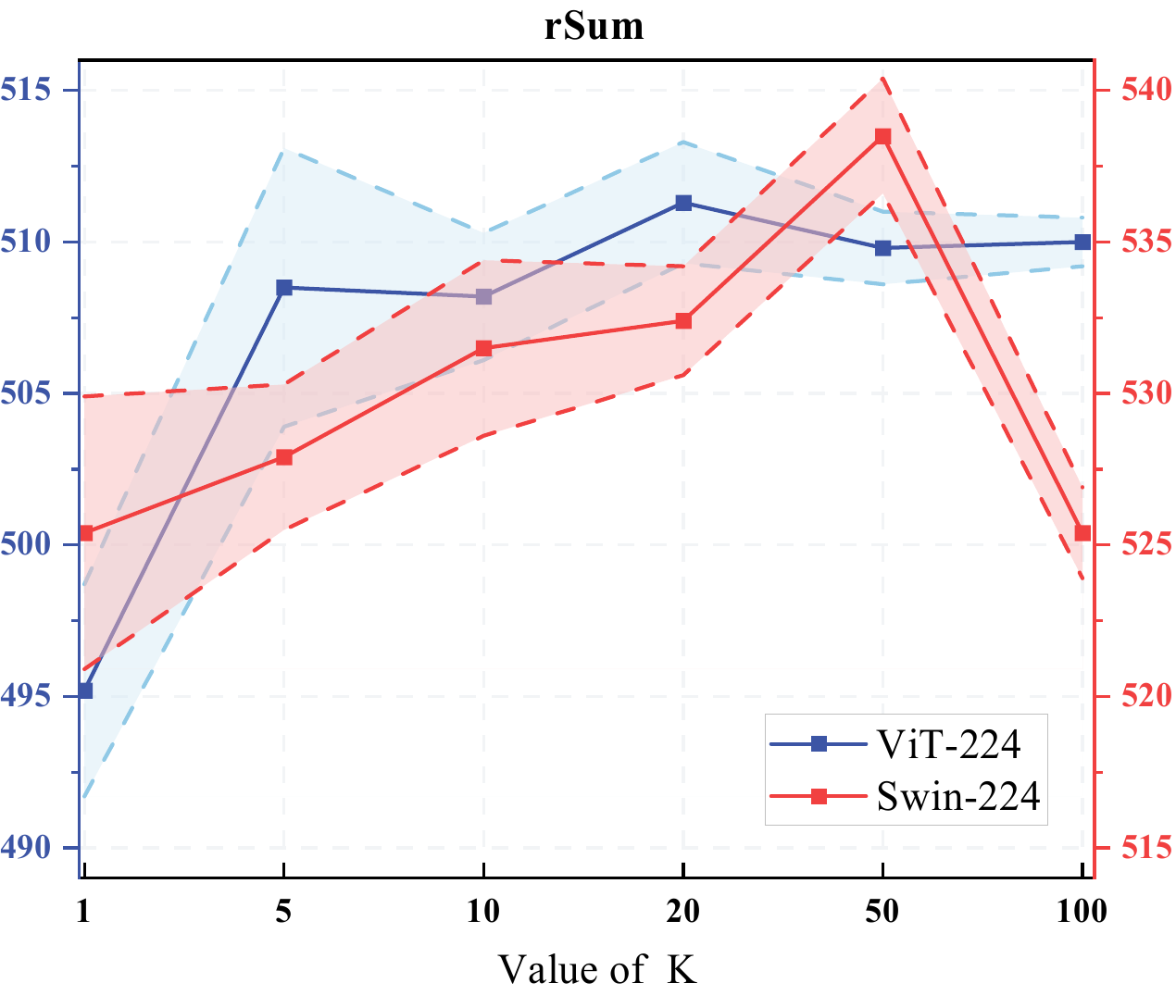}
	\includegraphics[width=0.33\linewidth]{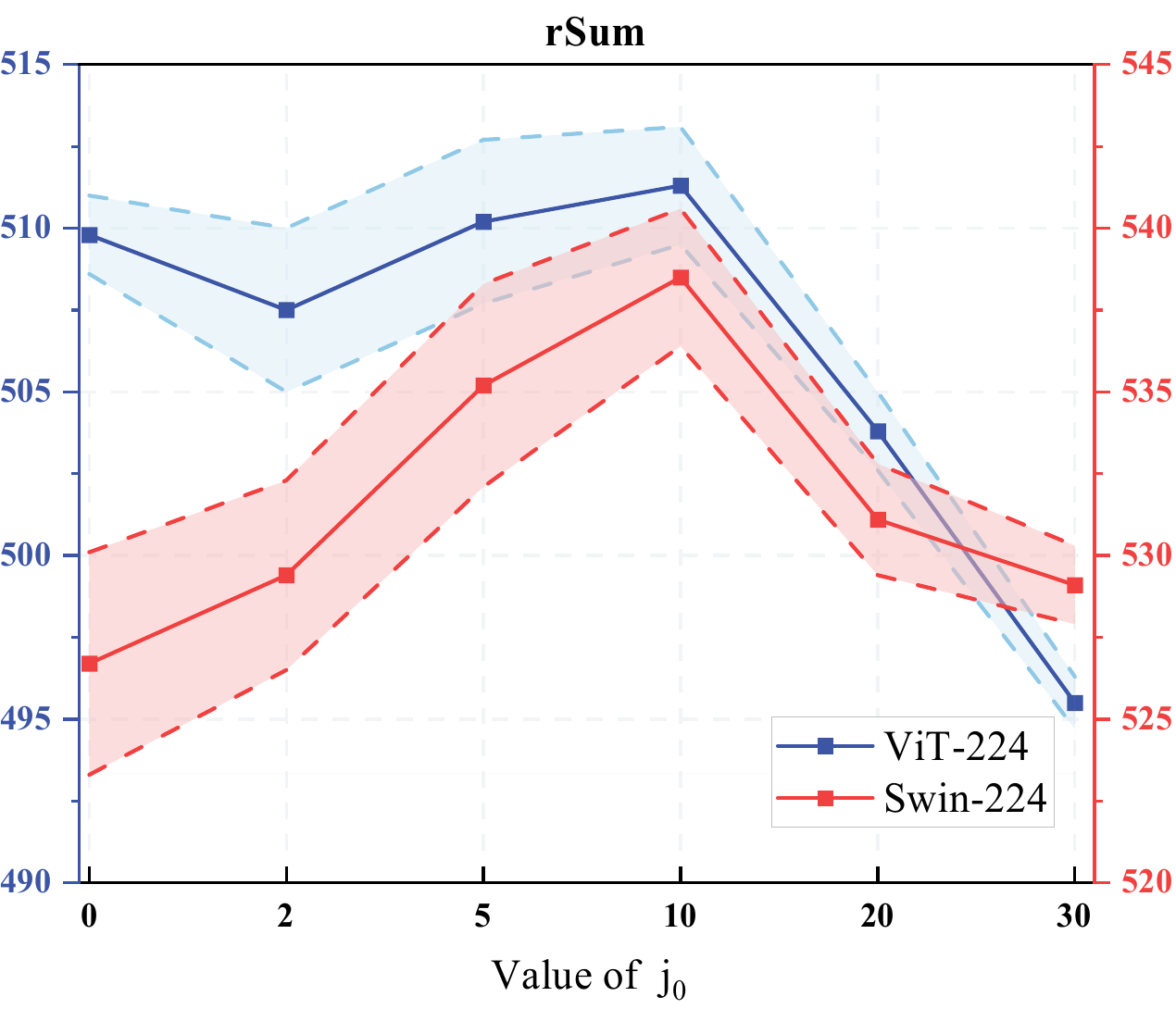}
	\includegraphics[width=0.33\linewidth]{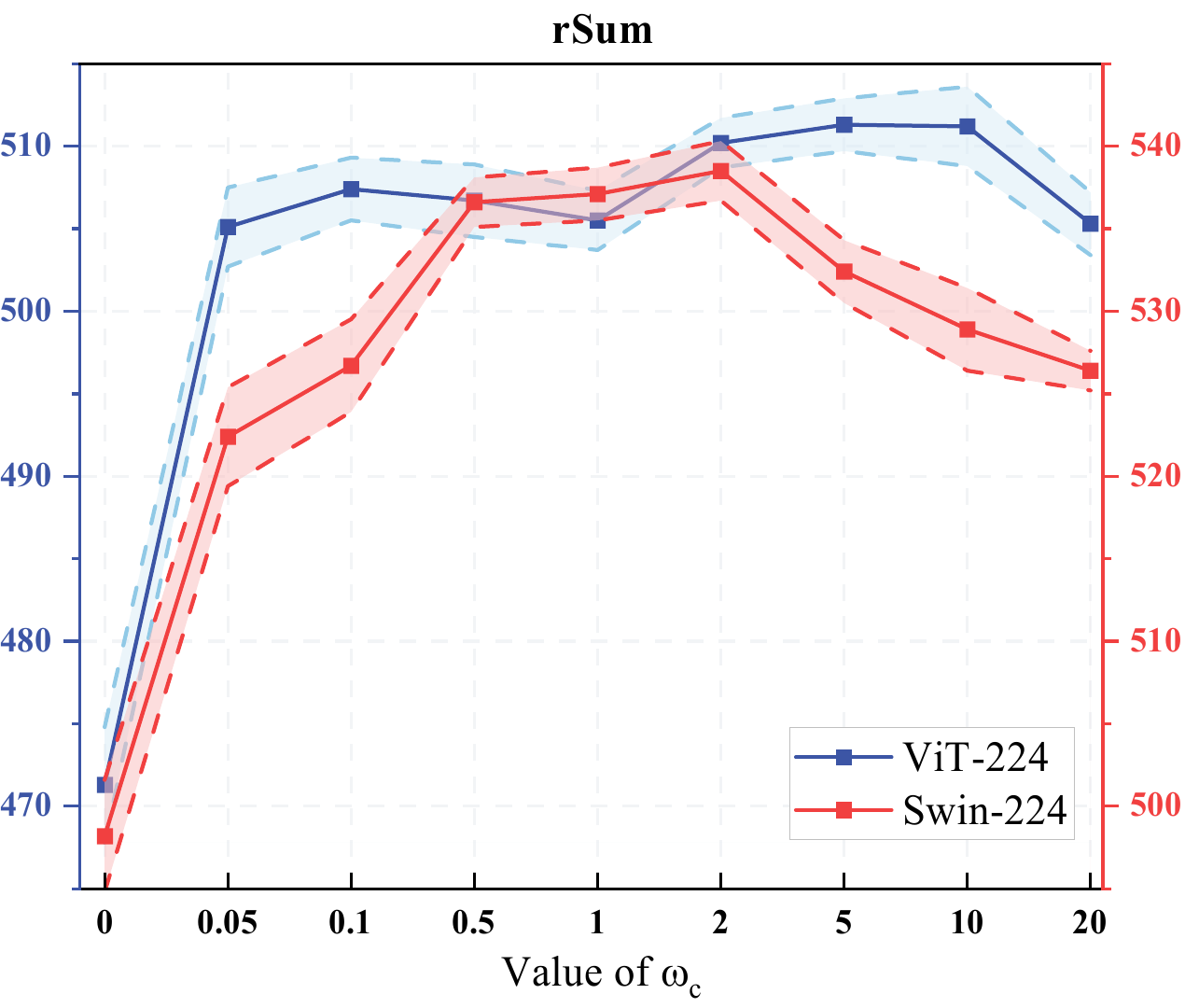}
	\caption{Performance comparison on varying hyperparameters.}
	\label{para}
\end{figure*}
To demonstrate the effectiveness of modules in PICO, we conduct ablation studies on both datasets, as shown in Tab.\ref{ab}. The baseline w/o $Wei$ means no weighting is applied to the embedding interaction. w/o $Pro$ means no prototype extraction is performed, using pseudo-semantic probability as weights. w/o $Ite$ means no iterative construction of prototypes is performed. w/o $Fed$ means no performance feedback-based weights. According to the experimental results, we have the following observations:

(1) \textbf{The effectiveness of model designing.} Removing any modules in PICO results in a performance decline, indicating that weighting the embedding interaction process is necessary, and the proposed pseudo-style prototype extraction, prototype iterative construction, and performance feedback-based weights can improve the overall performance of the model. 

(2) \textbf{Discussion on semantic probability calculation.} The results of w/o $Wei$ demonstrate that weighting feature columns in embedded interactions can significantly improve performance of model. There results of w/o $Pro$ indicate that the pseudo-semantic probability has been able to improve model performance, but the semantic probability constructed later is more effective.

(3) \textbf{Discussion on prototype iterative construction.} The performance of w/o $Ite$ is better than w/o $Pro$, indicating that our proposed prototype iterative construction can avoid performance degradation caused by independent clustering in different epochs. The results of w/o $Fed$ verify the effectiveness of performance feedback-based weights quantitatively. 

To further discuss the robustness of PICO, we apply it to other methods. The results are shown in Tab.\ref{cross}, which demonstrate the adaptive weighting for feature columns can also improve the performance of other models.
\begin{table}
	\centering
	\caption{Generalization ability comparison of models trained on MS-COCO and evaluated on Flickr30K.}
	\resizebox{0.48\textwidth}{!}{\begin{tabular}{l|ccc|ccc|c}
			\toprule
			& \multicolumn{3}{c}{Image-to-Text}& \multicolumn{3}{c|}{Text-to-Image}&\multirow{2}{*}{rSum}\\
			&R@1&R@5&R@10&R@1&R@5&R@10&\\
			\midrule
			\multicolumn{8}{l}{\textbf{\textit{ViT-224 + BERT}, 14$\times$14 patches}}\\
			Baseline&58.3&83.4&89.0&44.9&74.6&82.8&433.0\\
			\textbf{PICO}&63.5&84.7&91.4&49.9&76.0&84.8&450.3\\
			\hline
			\multicolumn{8}{l}{\textbf{\textit{Swin-224 + BERT}, 7$\times$7 patches}}\\
			Baseline&65.2&85.6&91.2&49.9&75.1&80.5&447.5\\
			\textbf{PICO}&68.7&88.1&92.5&54.5&82.9&85.1&471.8\\
			\bottomrule
	\end{tabular}}
	\label{gen}
\end{table}
\subsection{Visualization}
\textbf{Distribution of correlation score.} Fig.\ref{viscas} shows the visualization of the change in distribution of correlation score with / without PICO's weighting process during embedding interaction. The weighting process pushes correlation scores toward the extreme values(near 0 or 1), simplifying match assessment. By suppresses the role of style information, PICO ensures that patches or words with equivalent meaning achieve consistent scores.

\textbf{Correspondence between blocks and words.} Fig.\ref{vis} shows the visualization of patches corresponding to words with similar meanings. Both the original method and PICO use the same hard-threshold for patch selection. The red box in the figure indicates the differences of selected patches. It can be seen that compared to the original method, PICO can significantly reduce the differences in patches selected from words with similar meanings. This verifies our model's ability to reduce alignment differences caused by different text expression styles.
\subsection{Robustness Analysis}
\textbf{Parameter sensitivity.} Fig.\ref{para} shows the performance of PICO by varing values of hyper-parameters, inlucding the number of clusters $K$, the epoch of prototype initialization $j_0$, and the adjustment parameter $\omega_c$. When varying any of these hyper-parameters, we fix others with default settings. The optimal values for $K$, $j_0$ and $\omega_c$ are 20, 10, and 5 with the ViT-224 backbone, and 50, 10, and 2 with the Swin-224 backbone.

\textbf{Generalization study.} To evaluate the generalization capability of PICO in learning latent semantics, we perform cross-validation experiments following \cite{zhang2023unlocking}. The model trained on MS-COCO is directly evaluated on Flickr30K in a zero-shot setting. The results shown in Tab.\ref{gen} demonstrate that PICO outperforms the baseline in generalization performance, confirming its effectiveness in capturing cross-modal latent semantics. 
\section{Conclusion}
In this paper, we propose a reliable cross-modal alignment method based on prototype iterative construction (PICO). PICO reduces the weights of feature columns dominated by style information during the embedding interaction, to avoid the information bias or feature loss. Our work focuses on ensuring the reliability of those weights, for which we propose an iterative construction mechanism for prototypes and a performance feedback based update strategy. Extensive experiments and analyses conducted on various benchmarks and backbones demonstrate the superiority and rationality of our method.
\begin{acks}
	The authors appreciate the financial support by the National Natural Science Foundation of China (NSFC) under Grant Number 92367202, the NSFC Joint Fund with Zhejiang Integration of Informatization and Industrialization under Key Project (Grant Number U22A2033), the Postdoctoral Fellowship Program of CPSF under Grant Number GZC20251643.
\end{acks}
\bibliographystyle{ACM-Reference-Format}
\bibliography{sample-base}

\end{document}